\documentclass[a4paper,fleqn]{cas-sc}

\usepackage[numbers]{natbib}

\def\tsc#1{\csdef{#1}{\textsc{\lowercase{#1}}\xspace}}
\tsc{WGM}
\tsc{QE}
\tsc{EP}
\tsc{PMS}
\tsc{BEC}
\tsc{DE}

\begin{document}
\let\WriteBookmarks\relax
\def\floatpagepagefraction{1}
\def\textpagefraction{.001}
\shorttitle{Curvature graph neural network}
\shortauthors{H. Li et~al.}

\title [mode = title]{Curvature Graph Neural Network}

\author[1]{Haifeng Li}
\author[2]{Jun Cao}
\author[3]{Jiawei Zhu}
\author[2]{Yu Liu}
\author[3]{Qing Zhu}
\author[4]{Guohua Wu}

\cormark[1]

\cortext[cor1]{Corresponding author}
\address[1]{School of Geosciences and Info-Physics, Central South University, Changsha 410083, China}
\address[2]{School of Earth and Space Sciences, Peking University, Beijing, China}
\address[3]{Faculty of Geosciences and Environmental Engineering, Southwest Jiaotong University, Chengdu, China}
\address[4]{School of Traffic and Transportation Engineering, Central South University, Changsha 410075, China}

\begin{abstract}
Graph neural networks (GNNs) have achieved great success in many graph-based tasks. Much work is dedicated to empowering GNNs with the adaptive locality ability, which enables measuring the importance of neighboring nodes to the target node by a node-specific mechanism. However, the current node-specific mechanisms are deficient in distinguishing the importance of nodes in the topology structure. We believe that the structural importance of neighboring nodes is closely related to the importance of them in aggregation. In this paper, we introduce discrete graph curvature (the Ricci curvature) to quantify the strength of structural connection of pairwise nodes. And we propose Curvature Graph Neural Network (CGNN), which effectively improves the adaptive locality ability of GNNs by leveraging the structural property of graph curvature. To improve the adaptability of curvature to various datasets, we explicitly transform curvature into the weights of neighboring nodes by the necessary Negative Curvature Processing Module and Curvature Normalization Module. Then, we conduct numerous experiments on various synthetic datasets and real-world datasets. The experimental results on synthetic datasets show that CGNN effectively exploits the topology structure information, and the performance is improved significantly. CGNN outperforms the baselines on 5 dense node classification benchmark datasets. This study deepens the understanding of how to utilize advanced topology information and assign the importance of neighboring nodes from the perspective of graph curvature, and encourages us to bridge the gap between graph theory and neural networks.
\end{abstract}




\maketitle

\section{Introduction}
Inspired by the great success of deep learning in the Euclidean domain, GNNs attempt to generalize neural networks to non-Euclidean domains, e.g., graph. With the ability to automatically extract low-dimensional representations of nodes or graphs, GNNs have been widely used in recommender systems \cite{wang2019kgat, fan2019graph}, biochemistry \cite{fout2017protein, do2019graph}, social networks \cite{qiu2018deepinf}.

The core idea of GNNs is how to aggregate the features of neighboring nodes to enrich the representations of target nodes. Mainly based on the homophily assumption \cite{mcpherson2001birds, grover2016node2vec}, GNNs strengthen the similarity of node representations belonging to the same class and the difference of node representations between different classes, and smooth the representation of nodes on local structures based on connection. Smoothness is an intrinsic property of GNNs \cite{li2018deeper}, which makes neighboring node representations similar and is beneficial for neighboring nodes classified into the same class \cite{deng2019batch}. Smoothing is a double-edged sword for GNNs: proper smoothing can generate high-quality, low-dimensional node representations that are beneficial to downstream tasks, while oversmoothing makes node representations indistinguishable and is harmful to downstream tasks. Oversmoothing is mainly caused by the edges connecting different classes of nodes in real-world datasets, which dilute the useful information of the target nodes by aggregation and cause the different class of node representations too similar \cite{chen2020measuring}. To alleviate oversmoothing, we need to assign more weights to the same class of nodes and less weights to the other nodes.

GNNs mainly measure the importance of neighboring nodes in three ways. The simplest way considers the neighboring nodes equally important, such as Neural FPs \cite{duvenaud2015convolutional}, GraphSAGE \cite{hamilton2017inductive} and GIN \cite{xu2018powerful}. The second way assesses the weight of neighboring nodes by utilizing node degrees. A typical GNN adopting this way is GCN \cite{kipf2016semi}, which assigns the weights of neighboring nodes inversely proportional to the degree of neighboring nodes. The GCN-derived TAGCN \cite{du2017topology} and SGC \cite{wu2019simplifying} also leverage node degrees. Both of the ways explicitly compute the weights of neighboring nodes, while some other models implicitly generate the weights to adapt to datasets in a data-driven manner. For example, GAT \cite{velivckovic2017graph} calculates the weights through a self-attention mechanism and CurvGN \cite{ye2019curvature} transforms Ricci curvature into the weights by the MultiLayer Perceptron (MLP).

\begin{figure}
	\centering
		\includegraphics[scale=.75]{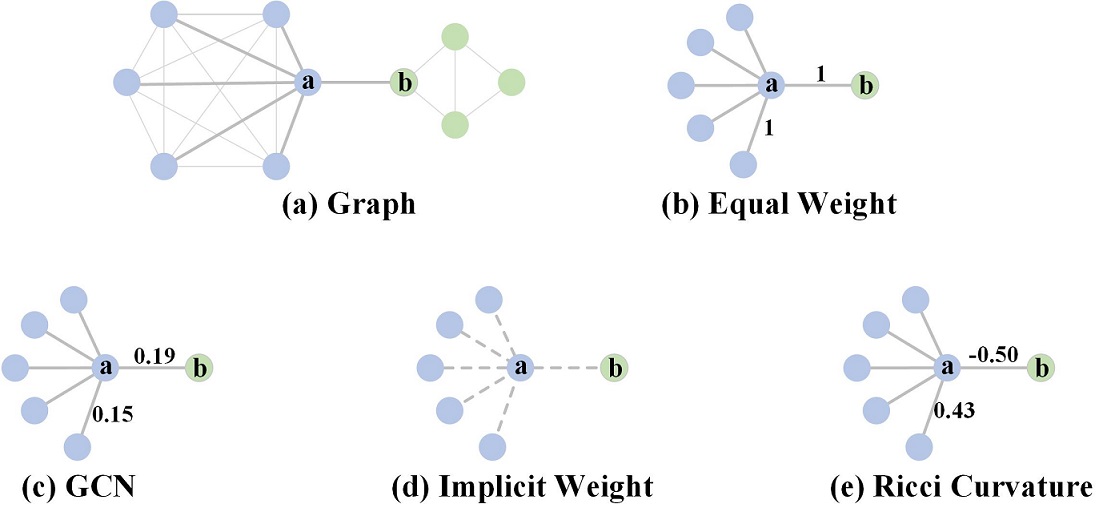}
	\caption{Illustration of different ways to compute the weights of neighbors of node a. (a) is the original graph which can be structurally divided into two classes: blue and green. For (b) and (c), the numbers of edges indicate the weights of neighbors in aggregation. The dashed lines of (d) represent the unknown weights. The numbers of (e) represent the Ricci curvature of the edges.}
	\label{fig: Weigh_Illu}
\end{figure}

However, the above three ways cannot properly characterize the structural importance of neighboring nodes. Many real-world datasets have an aggregating tendency for nodes, forming locally tight-connected groups with relatively sparse inter-group connection \cite{porter2009communities, girvan2002community}, named as communities. In general, nodes in the same community have strong similarities, while nodes between different communities have strong differences. Further, we argue that the more the overlap of neighboring nodes of a node pair, the stronger the structural connection between them, and vice versa. The nodes in Figure \ref{fig: Weigh_Illu}(a) are divided into two classes based on the structure. The key of aggregating the neighboring nodes for node $a$ is to weaken the influence of node $b$. Equal importance treats node $b$ and blue nodes as equivalents. Since the node degree of node $b$ is smaller than blue nodes, GCN considers it important. Implicit manners cannot measure the influence of node $b$ before finishing training. Due to utilizing limited topological information such as node degree or connection, all of these ways cannot directly or explicitly weaken the influence of node b. The above three ways are respectively shown by Figure \ref{fig: Weigh_Illu}(b), (c), (d).

\begin{figure}
	\centering
		\includegraphics[scale=.75]{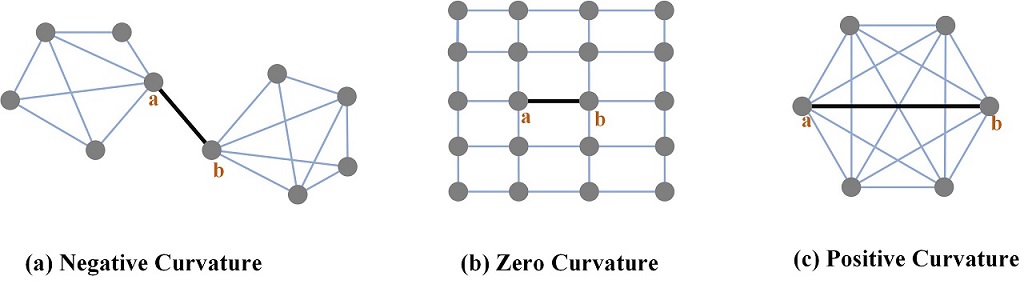}
	\caption{Illustration of structural information expressed by different values of the Ricci curvatures. The weights of all edges are 1.}
	\label{fig: Ricci_illu}
\end{figure}

In this paper, we introduce the concept of graph curvature to improve the local structural adaptability of GNNs. Graph curvature well measures how neighbors of a pair of nodes relate to each other \cite{lin2011ricci}. In analogy to curvature quantifying the deviation of a curve from a straight line in Euclidean space, discrete graph curvature measures the geometric deviation of the neighboring nodes of two nodes on an edge from a "flat" shape, e.g., the shape of a grid graph in which all nodes are structurally equivalent. There are several definitions of graph curvature, among which Ollivier’s Ricci curvature \cite{ollivier2009ricci, ollivier2010survey} is the most attractive. Ollivier’s Ricci curvature can quantify the strength of interaction or overlap between neighbors of a pair of nodes. In a (infinite) grid graph, the Ricci curvature of all edges is zero, as shown in Figure \ref{fig: Ricci_illu}(b). When the connection of nodes in Figure \ref{fig: Ricci_illu}(c) is denser, the Ricci curvature is positive. The edge $(a, b)$ in Figure \ref{fig: Ricci_illu}(a) connecting two independent groups which is like a “bridge” has a negative Ricci curvature. Naturally, the Ricci curvature diminishes the effect of node $b$ on node $a$ in Figure \ref{fig: Weigh_Illu}(a). And Figure \ref{fig: Weigh_Illu}(e) shows that the Ricci curvature of $(a, b)$ is -0.50, which is much smaller than the curvature between the blue node and node $a$. We argue that the Ricci curvature characterizes the relationship of the pairwise structure connections and should be exploited by GNNs.

Here, we propose Curvature Graph Neural Network (CGNN), which improves the discrimination ability for local structures by leveraging the Ricci curvature. For CurvGN, Li et al. \cite{li2021graph} argues that its success is due to the learning ability of MLP, not the Ricci curvature. To illustrate it, the performance of CurvGN does not change significantly even if we replace the Ricci curvature with random values randomly sampling from a uniform distribution of 0 to 1, as shown in Table \ref{tab: CurvGN with different values}. Unlike CurvGN which implicitly assign weights to different channels of node features by MLP, we explicitly transform the Ricci curvature into the weights of neighboring nodes. And the structural information represented by the Ricci curvature is effectively retained and utilized by CGNN. Since using the Ricci curvature as the weights makes CGNN difficult to train and degrades the performance, we propose the Negative Curvature Transformation Module (NCTM) and the Curvature Normalization Module (CNM), both of which do not destroy the relative magnitude of curvature, i.e., edges with large curvature have relatively large weights.

\begin{table}[width=.9\linewidth,cols=4,pos=h]
\caption{The mean accuracies of CurvGN with different information on Cora and PubMed. The column of Ricci Curvature means the performance of CurvGN using the Ricci curvature as the input of MLP, while the other columns indicate replacing the Ricci curvature with the random values by sampling from 0-1 uniform distribution with random seed 0, 10, 100 respectively.}
\label{tab: CurvGN with different values}
\begin{tabular*}{\tblwidth}{@{} LCCCC@{} }
\toprule
         & Ricci Curvature & Seed=0 & Seed=10 & Seed=100 \\
\midrule
        Cora & 82.3\% & 82.2\% & 82.3\% & 82.1\% \\
        PubMed & 78.9\% & 78.8\% & 78.9\% & 79.1\% \\
\bottomrule
\end{tabular*}
\end{table}

We conduct extensive experiments on various synthetic and real-world datasets to illustrate that the Ricci curvature improves the local structural adaptability of GNNs. The experimental results indicate that the Ricci curvature is beneficial for measuring the strength of pairwise structural connections and the performance of CGNN outperforms the baselines. Further, we visually express that CGNN can better weaken the interactions of node features between different classes. Finally, the ablation experiments prove the necessity of NCTM and CNM to guarantee the performance of CGNN. Our contributions are mainly as follows:
\begin{itemize} 
\item We propose CGNN, which significantly improves the local structural adaptability and the quality of node representations by transforming the Ricci curvature into the weights of neighbors; 
\item We propose NCTM to process negative Ricci curvature and CNM to utilize the structural information of nodes in different order hops, and both of them effectively help the Ricci curvature adapt to various datasets; 
\item CGNN outperforms the baselines on 5 dense node classification benchmark datasets
\end{itemize}

The paper is organized as follows: in Section 2, we briefly review the related work on GNNs and the discrete Ricci curvature. Section 3 describes the network architecture of CGNN and the pipeline of transforming the Ricci curvature. Section 4 evaluates the performance of CGNN on various synthetic and real-world datasets. Section 5 concludes this paper.

\section{Related work}
\textbf{\textit{Graph Neural Network.}} We can classify graph neural networks into two categories: spectral GNNs and spatial GNNs \cite{wu2020comprehensive}. Spectral GNNs define convolution through the eigendecomposition of the graph Laplacian matrix in the Fourier domain. Bruna et al. explored the generalization of CNNs to graph-structured data by the decomposition of the graph Laplacian matrix \cite{bruna2014spectral}. Although the matrix decomposition requires lots of computational resources, it does not filter the local information. This work attempts to reduce computation and make convolution spatially localized through a Graph Estimation procedure \cite{henaff2015deep}. Defferrard et al. approximate convolution with  truncated Chebyshev polynomials, avoiding the computation of the eigenvectors of the graph Laplacian matrix \cite{defferrard2016convolutional}. Kipf et al. proposed Graph Convolution Network (GCN) \cite{kipf2016semi}, which further simplifies convolution by setting the order of the truncated Chebyshev polynomials to 1 and a renormalization trick. GCN can still be simplified to a linear model \cite{wu2019simplifying} through successively removing nonlinear activation function and collapsing weight matrices between consecutive layers, which fits large-scale datasets and improves interpretability. Since the spectrum is inherently graph-specific, a serious problem for spectral GNNs is that the trained model cannot be generalized to other datasets.

Spatial GNNs directly define convolution on the local structure of the graph to improve the generalization. The key of Spatial GNNs is how to handle neighbors of different sizes and maintain the local invariance of convolution. A straightforward idea is to use different weight matrices for neighbors with different degrees \cite{duvenaud2015convolutional}, but this approach cannot be applied to large-scale graphs with rich node degrees. Monti et al. proposed MoNet \cite{monti2017geometric}, which directly generalizes CNNs to graphs and can automatically extract local, stationary, and compositional node representations. And some work \cite{li2015gated, tai2015improved} exploited Gate mechanisms such as GRU \cite{cho2014learning} and LSTM \cite{hochreiter1997long} to improve the ability to capture long-range structural information. To aggregate the features of neighbors, Hamilton et al. proposed GraphSAGE \cite{hamilton2017inductive}, which introduced a fixed-size neighbor sampling method and three different ways to aggregate neighbors. GAT \cite{velivckovic2017graph} introduced a self-attention mechanism into GNNs that implicitly assigns weights to neighbors. CurvGN \cite{ye2019curvature} used the Ricci curvature as additional structural knowledge and implicitly assigned specific weights to channels of neighboring node features by MLP. Recently, some work has attempted to map node features from Euclidean space into hyperbolic space \cite{chami2019hyperbolic}, spherical space \cite{defferrard2019deepsphere}, and constant curvature space \cite{bachmann2020constant}, in order to adapt to complex and variable real-world datasets. Inspired by self-supervised learning, there has also been a great deal of interest in how to facilitate GNNs through diverse and valuable graph information \cite{you2020does, jin2020self, qiu2020gcc, you2020graph}.

\textbf{\textit{Discrete Ricci Curvature.}} The Ricci curvature is defined on the manifold through the geodesics, so it can be generalized to the discrete setting. Many works have been devoted to extending the Ricci curvature on graphs, such as the Ollivier’s Ricci curvature based on optimal transport theory \cite{ollivier2009ricci, ollivier2010survey} and the Forman curvature based on the graph Laplacian \cite{forman2003bochner}. Forman curvature is widely used in large-scale network analysis \cite{weber2017characterizing} and image processing \cite{saucan2009combinatorial}. The Ollivier's Ricci curvature precisely measures the sparsity/denseness of connections on local structure, it is widely applied to detecting network backbone and congestion \cite{ni2015ricci, wang2014wireless}, studying financial market fragility \cite{sandhu2016ricci}, and detecting the community structure of network \cite{ni2019community, sia2019ollivier}. Although Ollivier’s Ricci curvature is harder to compute, it is more geometric than the Forman curvature. Therefore, we select the Ollivier’s Ricci curvature as the advanced graph information to improve the adaptive locality ability of GNN. Ni et al. systematically compares and analyzes the difference between these two types of Ricci curvature \cite{ni2019community}. Note that CurvGN has introduced Ricci curvature into GNNs, but the way it is introduced erases the unique property of the Ricci curvature, and is completely different from ours.

\section{Methodology}
In this section, we elaborate on the architecture of CGNN and represent how to transform the Ricci curvature into the weights of neighbors in the aggregation. Specifically, we first introduce the node classification task and formulate the forward propagation of CGNN according to the Message Passing Neural Network (MPNN) framework. Then, we describe the definition of Ollivier's Ricci Curvature in detail. Finally, we propose the Negative Curvature Transformation Module and Curvature Normalization module to make it better to enhance the local structural adaptability of GNNs.

\begin{figure}
	\centering
		\includegraphics[scale=.75]{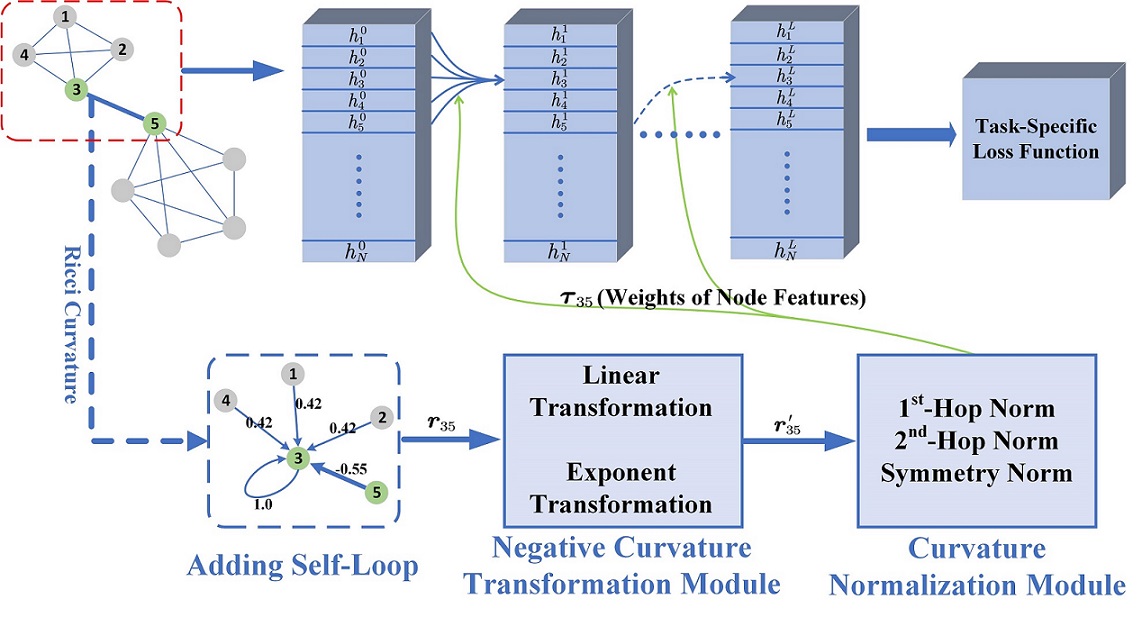}
	\caption{The architecture of the Curvature Graph Neural Network. The upper part shows message aggregation, and the lower part indicates the processing flow of the Ricci curvature.}
	\label{fig: Framework}
\end{figure}

\subsection{Curvature Graph Neural Network}
We evaluate the performance of different GNNs by node classification tasks. Firstly, the task is formulated. Suppose we have a graph $\mathcal{G}=(\mathcal{V}, \mathcal{E})$, with $N$ nodes $i \in \mathcal{V}$, $E$ edges $e_{i j}=(i, j) \in \mathcal{E}$ and node features $H=\left(h_{1}, h_{2}, \ldots, h_{N}\right)^{T} \in \mathbb{R}^{N \times F}$. Here, $h_{i}$ indicates the feature of node $i$, and $F$ is the dimension of $h_{i}$.Given the labels of some of the nodes, we aim at predicting the labels of the remaining nodes as accurately as possible by utilizing node features and topology structure informaion.

The MPNN framework is currently a mainstream framework of GNNs, which is both flexible and efficient. It consists of a message-passing part and a readout part. Since the readout is for the whole graph, we just focus on the message-passing part for node classification. Message is an alternative expression of the node feature. To maintain consistency of expression, we still use node features to represent messages. The message-passing part extracts localized information by iteratively transforming, aggregating, and updating node features, as shown in the upper part of Figure \ref{fig: Framework}. For some GNNs, the message-passing part can be summarized as Eq. \ref{eq: MPNN}:

\begin{equation}
\label{eq: MPNN}
h_{i}^{l+1}=\sigma\left(\square_{j \in \overline{\mathcal{N}}(i)}\left(\tau_{i j}^{l+1} W h_{j}^{l}\right)\right)
\end{equation}
where $\overline{\mathcal{N}}(i)=\mathcal{N}(i) \cup\{i\}$ indicates the neighboring nodes of the node $i$ after adding the self-loop, while $\tau_{i j}^{l+1}$ indicates the weight of node $j$ for node $i$ in the $l+1$ layer, and $\square$ represents a differentiable, permutation invariance aggregation function, e.g. mean, sum or max. We implement Curvature Graph Neural Network (CGNN) under MPNN framework.

Next, we describe the architecture of CGNN and the pipeline of transforming the curvature of edges into the weights of node features. Figure \ref{fig: Framework} illustrates the network architecture of CGNN, which includes a layer-stacked aggregation part and generating the weights of node features. To aggregate the features of neighboring nodes, we choose summation as the aggregation function, which is also widely adopted by other GNNs. Similar to Eq. \ref{eq: MPNN}, the forward propagation of CGNN is Eq. \ref{eq: CGNN}:

\begin{equation}
\label{eq: CGNN}
h_{i}^{l+1}=\sigma\left(\Sigma_{j \in \overline{\mathcal{N}}(i)}\left(\tau_{i j} W h_{j}^{l}\right)\right)
\end{equation}
where $\tau$ denotes the weights of neighboring nodes based on the curvature which is fixed in any layer. Here, we choose Ollivier-Ricci Curvature as the curvature utilized in CGNN because of being more geometrical. To generate the weights, we first need to add self-loops to nodes, then ensure that the curvature is positive by the negative curvature processing module, and finally normalize it. The workflow of the Ricci curvature is shown in the lower part of \ref{fig: Framework}. Same as GCN, we set layers of CGNN to 2.

\subsection{Ollivier’s Ricci Curvature}
Curvature can quantitatively measure the degree of deviation in space. In Euclidean space, curvature measures the degree to which a curve deviates from a straight line, or a surface deviates from a plane. In Riemannian geometry, curvature measures the degree to which the manifold deviates from the Euclidean space, and the Ricci curvature quantifies its deviation in the orthogonal direction. And the Ricci curvature determines the rate at which the volume of a ball grows, expressed as a function of the radius. It also determines the overlapping volume of two balls and the distance between the centers of them. If the overlapping volume is larger, the cost of the transfer is less, which indicates that the Ricci curvature and optimal transport theory are closely related. Ollivier bridges the gap between them and generalizes the Ricci curvature to the metric space by optimal transport theory.

We now introduce a key distance function of optimal transport theory: Wasserstein distance. Given a metric space $(M,d)$, $m_{x}$ and $m_{y}$ are the two probability measures with mass 1 respectively. Wasserstein distance $W\left(m_{x}, m_{y}\right)$ means the minimum average mass-preserving transportation plan between $m_{x}$ and $m_{y}$. Then, the coarse Ricci curvature (the Ricci curvature of Markov chains on metric spaces) generalizes the Ricci curvature to discrete spaces called Ollivier’s Ricci curvature. For simplicity, we still use the Ricci curvature to denote Ollivier's Ricci curvature. For graph, we define a probability distribution $m_{i}$ for each node $i$ . The Ricci curvature of an edge $e_{ij}$ can be formulated as Eq. \ref{eq: ricci curvature}:

\begin{equation}
\label{eq: ricci curvature}
r_{i j}=1-\frac{W\left(m_{i}, m_{j}\right)}{d(i, j)}
\end{equation}
where $d(i, j)$ indicate the shortest distance between node $i$ and $j$. We choose a simple and effective probability distribution with a hyperparameter $\alpha$ , as in (Ricci curvature of graphs). For an undirected and unweighted graph, the probability distribution of the node $i$ with node degree $k$ is Eq. \ref{eq: node distribution}:

\begin{equation}
\label{eq: node distribution}
m_{i}=\left\{\begin{array}{cl}
\alpha & \text { if } j=i \\
(1-\alpha) / k & \text { if } j \in \mathcal{N}(i) \\
0 & \text { otherwise }
\end{array}\right.
\end{equation}
Following the existing work \cite{ni2015ricci}, we set $\alpha=0.5$. Besides, the Ricci Curvature can be easily generalized to directed or weighted graphs. And see \cite{lin2011ricci} for more details. We set the Ricci curvature of the self-loop to 1.

The Ricci curvature incorporates rich topology structure information in terms of the graph theory. If the curvature is positive, i.e., $W\left(m_{x}, m_{y}\right)$ being smaller than $d(i, j)$, it indicates that the neighboring nodes of two nodes trend towards aggregation. It also implies that these two nodes are relatively closely related in structure. If the curvature of most of the edges is positive in the local structure, it can usually be referred to as Community \cite{sia2019ollivier}. On the contrary, the neighboring nodes of two nodes trend towards separation when the curvature is negative.

\subsection{Negative Curvature Transformation Module}
Negative curvature seriously degrades the performance of CGNN. Note that there are always some edges in graphs of which Ricci curvature is negative, such as $e_{ab}$  in Figure \ref{fig: Ricci_illu}(a). And most of the edges are with negative Ricci curvature when nodes are sparsely connected in the local structure. If we directly use the Ricci curvature including negative curvature as the weights of node features in aggregation, the negative curvature would make CGNN difficult to train. Besides, the negative curvature also has a serious impact on the symmetric normalization in the Curvature Normalization Module. Node degrees $d_{i}=\sum_{j \in \overline{\mathcal{N}}(i)} r_{i j}$ (the Ricci curvature being the weights of the edges) may be negative, while the symmetric normalization requires calculating the root of node degrees and may result in the imaginary weights of node features.

To solve the dissatisfied adverse effect of negative curvature, we propose the Negative Curvature Transformation Module (NCTM) to convert the negative curvature into positive numbers. We first explored the linear transformation, in which all of the Ricci curvature simply subtracts the minimum curvature and adds a positive number. We formulate the linear transformation as Eq. \ref{eq: linear transformation}:

\begin{equation}
\label{eq: linear transformation}
r_{i j}^{\prime}=r_{i j}-\min _{m, n}\left(r_{m n}\right)+\epsilon
\end{equation}
where $\epsilon \geqslant 0$ denotes the value corresponding to the minimum curvature. The linear transformation is intuitive and efficient, and ensures that the difference of curvature between edges is constant.

We also try to augment the difference between positive and negative curvature by exponential transformation, since positive and negative curvature indicate distinctly different signatures. We choose the sigmoid function to implement the transformation. The formula of the sigmoid function is Eq. \ref{eq: exp transformation}:

\begin{equation}
\label{eq: exp transformation}
r_{i j}^{\prime}=\frac{1}{1+e^{-r_{i j}}}
\end{equation}
The sigmoid function, as a common single increasing function, can not only transform negative curvature into positive numbers, but also enlarge the difference between positive and negative curvature by concentrating positive curvature around 1 and negative curvature around 0.

The NCTM ensures that there is no negative curvature after data processing. Both of the linear transformation and the exponential transformation do not destroy the relative magnitude relationship of curvature, and effectively preserve the property of the Ricci curvature.

\subsection{Curvature Normalization Module}
GCN smooths the features of neighboring nodes by normalizing node degrees. Li et al. \cite{li2018deeper} consider the convolution of GCN as a special kind of the Laplacian smoothing. The Laplacian smoothing can be formulated as Eq. \ref{eq: Laplacian smoothing}:

\begin{equation}
\label{eq: Laplacian smoothing}
Y=\left(I-\gamma \tilde{D}^{-1} \tilde{L}\right) X
\end{equation}
where $\tilde{A}=A+I$, $\tilde{d}_{i i}=\sum_{i=1}^{N} \tilde{a}_{i j}$, $\tilde{L}=\tilde{D}-\tilde{A}$ and $0<\gamma \ll 1$ is a parameter balancing the target node and neighboring nodes. By setting $\gamma=1$, we get $Y=\tilde{D}^{-1} \tilde{A} X$, , which is the standard form of Laplacian smoothing. Then, we get $Y=\tilde{D}^{-1 / 2} \tilde{A} \tilde{D}^{-1 / 2} X$ by replacing the regularized Laplacian matrix $\tilde{D}^{-1} \tilde{L}$ with the symmetric regularized Laplacian matrix $\tilde{D}^{-1 / 2} \tilde{L} \tilde{D}^{-1 / 2}$, which is the graph convolution of GCN.

We rethink the Laplacian smoothing from a structural perspective. The regularized Laplacian smoothing $\tilde{D}^{-1} \tilde{A} X$ can be considered as aggregating node features according to normalizing the degrees of $1^{st}$-hop nodes, and is called the $1^{st}$-hop normalization. Besides, we also asset the effect of $2^{nd}$-hop nodes for normalizing the degrees, which can be formulated as $\tilde{A} \tilde{D}^{-1} X$. The $2^{nd}$-hop normalization views the importance of neighboring nodes to the target node as inversely proportional to the degree of the neighboring nodes. It means that the more neighbors a neighboring node has, the less important it is to the target node, and vice versa. For example, a person with many friends in a social network is inevitably energy-distracted from maintaining relationships and weakens his influence on  friends, while a person with only a few friends will focus more on managing friendships and therefore increase his influence. The symmetric regularized Laplacian smoothing $\tilde{D}^{-1 / 2} \tilde{A} \tilde{D}^{-1 / 2} X$ seems to balance the structural information of $1^{st}$-hop and $2^{nd}$-hop nodes.

Here, we utilize the structural information of different orders of hop to normalize the Ricci curvature. We take the curvature processed by NCTM as the weights of the edges, which is obtained as the adjacency matrix $R^{\prime}$ and the corresponding degree matrix $D^{\prime}$. We further use the above three normalization ways to calculate the weights of node features in aggregation. To be consistent with the MPNN framework, we reformulate the three normalization ways to the node level. The formula of the $1^{st}$-hop normalization is Eq. \ref{eq: 1-hop norm}:
\begin{equation}
\label{eq: 1-hop norm}
\tau_{ij}=\frac{r_{i j}^{\prime}}{d_{i i}^{\prime}}
\end{equation}
The formula of the $2^{nd}$-hop normalization is Eq. \ref{eq: 2-hop norm}:
\begin{equation}
\label{eq: 2-hop norm}
\tau_{ij}=\frac{r_{i j}^{\prime}}{d_{j j}^{\prime}}
\end{equation}
The formula of the symmetric normalization is Eq. \ref{eq: symmetric norm}:
\begin{equation}
\label{eq: symmetric norm}
\tau_{i j}=\frac{r_{i j}^{\prime}}{\sqrt{d_{i i}^{\prime} \cdot d_{j j}^{\prime}}}
\end{equation}
The results in Section 4.5 show that $1^{st}$-hop normalization is appropriate on some datasets, while $2^{nd}$-hop normalization is more suitable for some other datasets. The performance of symmetric normalization always seems to be a trade-off between $1^{st}$-hop normalization and $2^{nd}$-hop normalization.

\section{Experiments and results}
\subsection{Data description}
To illustrate the importance of Ricci curvature for Message-passing, we respectively conducted extensive experiments on synthetic and real-world datasets with diversiform structures. We first utilize the Stochastic Block Model (SBM) \cite{abbe2017community} to generate synthetic graphs with community structure. SBM can custom the intra-community/inter-community probability for random-sampling edges. In contrast, we select the Erdős–Rényi model \cite{erdHos2013spectral} as a null model where all of the nodes are randomly connected with the same probability. Besides, we also explore the hub structure that a few nodes have significantly larger degrees with the Barabási–Albert model \cite{albert2002statistical}. For the real-world dataset, we analyze undirected and directed graphs in detail, and Table 2 summaries statistical details of the datasets. Cora, Citeseer, PubMed, Coauthor CS, and Coauthor Physics are citation networks, while Amazon Computers and Amazon Photo are e-commerce networks \cite{shchur2018pitfalls}. WikiCS \cite{mernyei2020wiki} is a directed graph from Wikipedia pages on the topic of computer science.

\begin{table}[width=.9\linewidth,cols=4,pos=h]
    \caption{Statistical details of all datasets.}
    \label{tab: dataset}
    \begin{tabular*}{\tblwidth}{@{} lccccccc@{} }
    \toprule
     & Nodes & Edges & Features & Classes & Training & Undirected & Avg Degree \\
    \midrule
        Cora             & 2078  & 5429  & 1433     & 7       & 140      & True       & 3.9 \\
        Citeseer         & 3327  & 4732  & 3703     & 6       & 120      & True       & 2.7 \\
        PubMed           & 19717 & 44338 & 500      & 3       & 60       & True       & 4.5 \\
        Coauthor CS      & 18333 & 10027 & 6805     & 15      & 300      & True       & 8.9 \\
        Coauthor Physics & 34493 & 282455& 8415     & 5       & 100      & True       & 14.4 \\
        Amazon Computers & 13381 & 259159& 757      & 10      & 200      & True       & 35.8 \\
        Amazon Photo     & 7487  & 126530& 745      & 8       & 160      & True       & 31.1 \\
        WikiCS           & 11701 & 216123& 300      & 10      & 580      & True       & 36.9 \\
    \bottomrule
    \end{tabular*}
\end{table}

\subsection{Baselines}
\subsubsection{Baselines for Synthetic dataset}

We selected four models for comparison on synthetic datasets for comparison, including three state-of-the-art models: GCN \cite{kipf2016semi}, GAT with concatenation \cite{velivckovic2017graph} and CurvGN \cite{ye2019curvature}. GCN uses the node degree to explicitly compute the weights of node features, while GAT and CurvGN implicitly generate the weights of node features in a data-driven manner. GAT introduces a self-attentive mechanism whose input is the node features of the hidden layer. CurvGN transforms the Ricci CurvGN into the multi-channel weights of node features through MLP. The remaining model is MLP, a particular kind of special GNNs which cannot take advantage of topology structure at all to generate node representations. 

\subsubsection{Baselines for Real-world dataset}

For real-world datasets, in addition to the above four models, we also compared Node2Vec \cite{grover2016node2vec}, MoNet \cite{monti2017geometric}, SAGE with mean aggregation \cite{hamilton2017inductive}, SGC \cite{wu2019simplifying} and APPNP \cite{klicpera2018predict}. These models were chosen as they utilized different graph structure information.

\subsection{Experiments setup}
\subsubsection{Dataset Division}

\textbf{Synthetic dataset.} Each dataset consisted of 1000 nodes, which are equally divided into 5 classes. We randomly select 20 nodes from each class as the training set, 300 nodes as the validation set, and the remaining 600 nodes as the test set. Then, we randomly generate a 20-dimensional feature for each node as node features. For SBM, we first generate 100 graphs of which intra-community probabilities $p$ of randomly sampling edges range in $\{0.05,0.07, \ldots, 0.23\}$ and inter-community probabilities $q$ range in $\{0.0,0.005, \ldots, 0.045\}$. And nodes within the same community are in the same class. For the dataset generated by Erdős–Rényi, the probability of randomly sample edges is 0.01.

\textbf{Real-world dataset.} For Cora, Citeseer, and PubMed, the datasets are divided in the same way as \cite{yang2016revisiting}. For Coauthor CS, Coauthor Physics, Amazon Computers, and Amazon Photo, the division is the same as \cite{shchur2018pitfalls}. Note that the division of the four datasets for CurvGN is different from \cite{shchur2018pitfalls}. WikiCS has a total of 20 kinds of division, each of which contains validation sets for hyperparameter selection and validation sets for early stopping respectively. We choose the first division for WikiCS, only utilizing the validation set for early stopping, and discard the validation set for hyperparameter selection.

\subsubsection{Hyperparameter setting}

To ensure the reproducibility of the paper, we choose 2020 as the random seed for any randomization operation. For the synthetic datasets, we set the dimension of the hidden layer in models to 8 and set the head of GAT to 1. For the real-world dataset, we adjust the hidden layer dimension of CGNNs to 64. And the architectures of baselines are consistent with their corresponding papers. We use Adam SGD optimizer with a learning rate of 0.005 and L2 regularization of 0.0005, and cross-entropy as the loss function to train the models. In this paper, the weight matrix is initialized with Glorot initialization. We use an early stopping strategy based on the validation set's accuracy with a apatience of 100 epochs. All models are trained on a single card Nvidia 2080Ti and are implemented via pytorch\_geometric \cite{Fey/Lenssen/2019}.

\subsubsection{Types of CGNN} 

For CGNN, we compared different configurations of NCTM and CNM. For example, CGNN\_Linear\_Sym means that we adopt the linear transformation and symmetric normalization operations, and CGNN\_Exp\_$1^{st}$ is the combination of exponential transformation and $1^{st}$-hop normalization. CGNN\_***\_$2^{nd}$ indicates CGNN with $1^{st}$-hop normalization. The rest is denoted similarly.

\subsubsection{Evaluation metric}

The evaluation metric of the paper is the mean and standard deviation of classification accuracy on test nodes. And we report the statistical results of 10 runs and 100 runs on synthetic datasets and real-world datasets respectively. For the sake of fairness, we re-use the metrics reported by \cite{monti2017geometric, shchur2018pitfalls, velivckovic2017graph} for some baselines on some real-world datasets.

\subsection{Effect of Different Topology Structure on GNNs}
\begin{figure}
	\centering
		\includegraphics[width=0.95\textwidth]{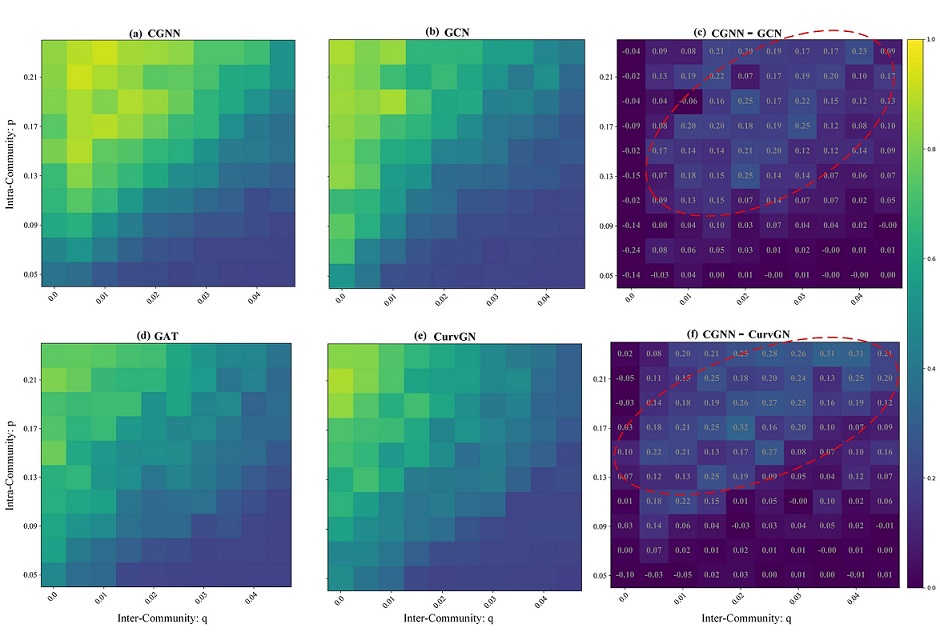}
	\caption{Heatmaps of accuracies on synthetic datasets for Stochastic Block Model. (a), (b), (d) and (e) represent the performance of CGNN, GCN, GAT and CurvGN respectively. (c) indicates the difference between CGNN and GCN, and (f) is between CGNN and CurvGN.}
	\label{fig: SBM_Comparision}
\end{figure}

To measure the effect of community structure on node feature aggregation, we investigated the classification accuracy of different models on 100 graphs generated by SBM, as shown in Figure \ref{fig: SBM_Comparision}(a), (b), (d), and (e). For all heatmaps, the classification accuracy tends to decrease from the top left to the bottom right. This trend indicates that the performance of GNNs is always better when is large and is small. Note that the smoothing of GNN enables it to distinguish nodes based on community structure, even if node features are randomly sampled. If the and are close, SBM gradually degrades into the null model Erdős–Rényi. And classification accuracies of models are around 20\%, which indicates that GNNs cannot classify nodes anymore. This result illustrates that appropriate smoothing is the key for GNNs, but too many inter-community edges dilute the useful information received by the target nodes, resulting in node features indistinguishable.

CGNN can better alleviate the effect caused by increasing inter-community edges. We subtract the accuracy of CGNN on 100 random graphs from GCN's and CurvGN's, as shown in \ref{fig: SBM_Comparision}(c), (f). We find that CGNN outperforms GCN and CurGN in the area of relatively large intra-community probabilities $p$ and relatively small inter-community probabilities $q$, i.e., the red circle. The structure of real-world datasets is often similar to the red circles with community structure. For such datasets, the Ricci curvature can appropriately increase the weights of intra-community nodes and decrease the weights of inter-community nodes according to the topology structure, improving the quality of aggregation.

\begin{figure}
	\centering
		\includegraphics[width=0.95\textwidth]{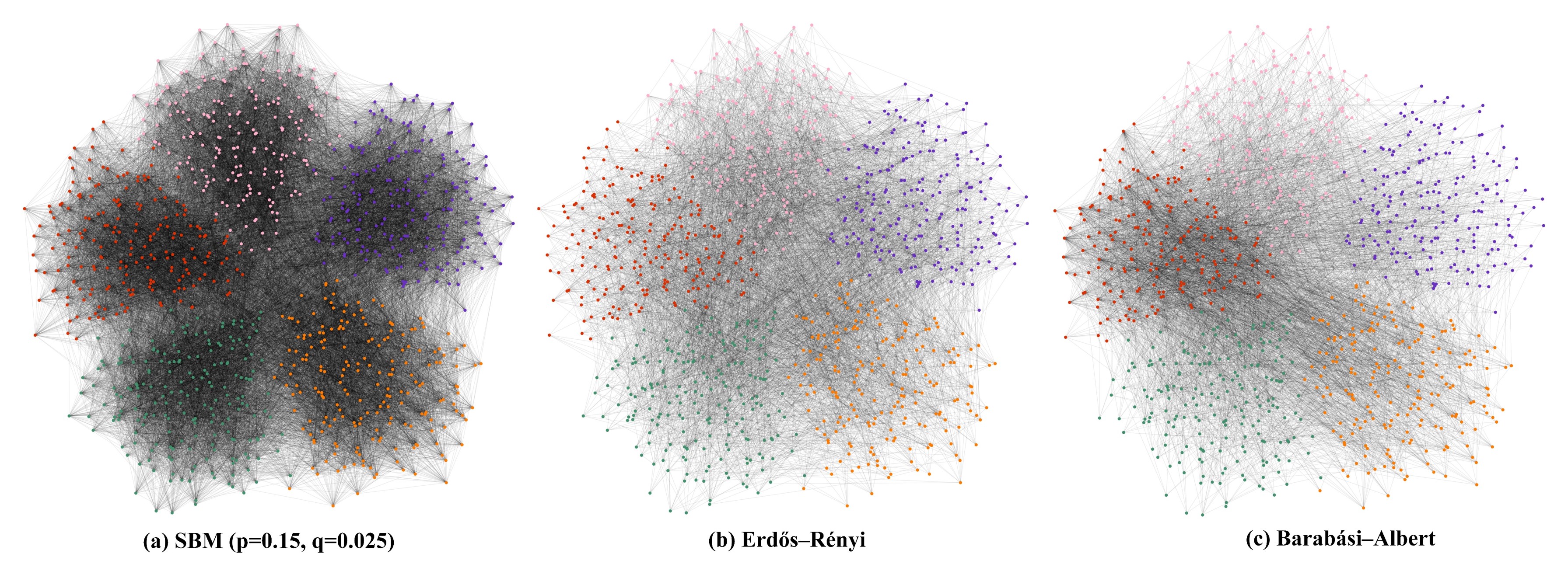}
	\caption{Visualization of three random graphs with different structures. The colors of nodes indicate the classes. In (a), the graph presents noticeable community structure. In (b), arbitrary pairs of nodes seem to be randomly connected. In (c), some of the red nodes are hubs.}
	\label{fig: Synthesis_Dataset_illu}
\end{figure}

We further explored the effect of different types of structures on GNNs. We use the graph of SBM $(p=0.15, q=0.025)$ as a complement to the real-world dataset with community structure, as shown in Figure \ref{fig: Synthesis_Dataset_illu}(a). And Figure \ref{fig: Synthesis_Dataset_illu}(b), (c) show the graphs generated by Erdős–Rényi and Barabási–Albert respectively. Table \ref{tab: result_synthetic_datasets} summaries the classification accuracy of different models on these three datasets. The classification accuracy of MLP is always around 20\%, which indicates that GNNs cannot discriminate nodes if discarding the structure information. The accuracy of CGNN is significantly higher than the other models on SBM. It illustrates that the Ricci curvature remarkably enhances the adaptive locality ability of GNNs. For Erdős–Rényi, the classification accuracy of models is around 20\%, which implies that the topology structure has a significant impact on the performance of GNNs. The accuracy of CGNN on Barabási–Albert is slightly higher than 20\%, but lower than that of GCN.The reason is that GCN can reduce the impact of hub nodes. However, the hub structure makes it possible for the curvature of the edges connecting different classes of nodes to be positive, incorrectly assign weights, and make node features indistinguishable, as the TexasChristian and UtahState nodes in Figure \ref{fig: Football}.

\begin{table}[width=.9\linewidth,cols=4,pos=h]
    \caption{Summary of statistical results in terms of classification accuracies (in percent) on different graph models.}
    \label{tab: result_synthetic_datasets}
\begin{tabular*}{\tblwidth}{@{} LCCCCC@{} }
\toprule
         & MLP          & GCN                   & GAT          & CurvGN       & CGNN \\
    \midrule
        SBM               & 19.3$\pm$2.1 & 43.1$\pm$11.0         & 48.6$\pm$6.5 & 35.9$\pm$7.5 & \textbf{62.1$\pm$8.2} \\
        Erd\"os-R\'enyi   & 19.7$\pm$1.2 & 19.9$\pm$1.5          & \textbf{19.9$\pm$1.2} & 19.6$\pm$1.4 & 19.7$\pm$1.2 \\
        Bar\'abasi-Albert & 19.5$\pm$1.6 & \textbf{34.4$\pm$2.5} & 20.5$\pm$1.6 & 20.5$\pm$1.5 & 22.9$\pm$ 2.3\\
\bottomrule
\end{tabular*}
\end{table}

\subsection{Node Classification on Benchmark Datasets}
In this section, we test the performance of CGNN on the node classification benchmark datasets. And the classification accuracies of CGNN and baselines are shown in Table \ref{tab: result_realworld_dataasets}. The accuracy of CurvGN on Coauthor CS, Coauthor Physics, Amazon Computers, and Amazon Photo differs from its original paper due to the fact that the paper of CurvGN divides these four datasets differently from \cite{shchur2018pitfalls}. Therefore, we retested the performance of CurvGN on all datasets. Note that the best results of CGNN are comparable to or even better than SOTA models. For the datasets with a small average node degree, the curvature of edges is mostly negative, which leads to the fact that the curvature does not characterize the importance of neighboring nodes very well. APPNP achieves the best accuracies in these datasets by exploiting a large and adjustable neighborhood by personalized PageRank. However, the performance of CGNN is always superior for datasets with relatively higher average node degrees. And the second-best results are mostly distributed in CGNN. Even on the directed dataset WikiCS, CGNN also achieves the best accuracy. These datasets with higher average node degrees have more nodes and edges, and their local structures are more diverse and complex. For the datasets with heterogeneous topology, the Ricci curvature can delicately and precisely measure the connection strength of pairwise nodes in terms of structure, and help CGNN better aggregate the features of neighboring nodes.

We further explored the impact of different combinations of NCTM and CNM on CGNN. For datasets with a small average node degree, exponential transformation can widen the gap between positive and negative curvature, thus helping to assist the aggregation of part of nodes. In this case, the exponential transformation may be slightly better than linear transformation. However, the exponential transformation for heterogeneous datasets with a larger average node degree would concentrate the curvature around 0 and 1 to weaken the hierarchy of it. The linear transformation is isometric and preserves the structural information of the curvature. The linear transformation can outperform the exponential transformation under the condition. In addition, we observe that the best results of CGNN are always distributed on $1^{st}$-hop normalization or $2^{nd}$-hop normalization, while the symmetric normalization seems to be the trade-off. The result implies that we can try to normalize the curvature with different orders of hub nodes to pursue better accuracy. It is worthwhile to investigate what kind of normalization is suitable for which property of datasets.

\begin{table}[]
    \caption{Summary of statistical results in terms of mean test set classification accuracies (in percent) and standard deviation on eight node classification benchmark datasets. \textbf{\textcolor{red}{Red}} numbers indicate the best performance, and bolded numbers mean the second best performance. OOM means out of memory.}
    \label{tab: result_realworld_dataasets}
    \setlength{\tabcolsep}{2mm}{
    \begin{tabular}{ccccccccc}
    \toprule
         & Cora & Citeseer & PubMed & CS & Physics & Computers & Photo & WikiCS \\
    \midrule
        MLP       & 59.0$\pm$0.9 & 58.9$\pm$0.7 & 67.1$\pm$0.5 & 88.9$\pm$0.6 & 87.5$\pm$0.8 & 67.7$\pm$1.2 & 81.8$\pm$0.8 & 72.5$\pm$0.2 \\
        Node2Vec  & 71.5$\pm$1.0 & 62.3$\pm$0.9 & 73.1$\pm$1.3 & 82.1$\pm$0.8 & 75.8$\pm$1.2 & 86.8$\pm$0.6 & 71.8$\pm$0.4 & 71.6$\pm$0.4 \\
        GCN       & 81.5$\pm$1.3 & 71.9$\pm$0.9 & 77.8$\pm$2.9 & 91.1$\pm$0.5 & 92.8$\pm$1.0 & 82.6$\pm$2.4 & 91.2$\pm$1.2 & 72.4$\pm$0.3 \\
        MoNet     & 81.3$\pm$1.3 & 71.2$\pm$2.0 & 78.6$\pm$2.3 & 90.8$\pm$0.6 & 92.5$\pm$0.9 & 83.5$\pm$2.2 & 91.2$\pm$1.3 & OOM \\
        GraphSAGE & 79.2$\pm$7.7 & 71.6$\pm$1.9 & 77.4$\pm$2.2 & 91.3$\pm$2.8 & 93.0$\pm$0.8 & 82.4$\pm$1.8 & 91.4$\pm$1.4 & 78.0$\pm$0.2 \\
        GAT       & 81.8$\pm$1.3 & 71.4$\pm$1.9 & 78.7$\pm$2.3 & 90.5$\pm$0.5 & 92.5$\pm$0.9 & 78.0$\pm$19.0 & 85.7$\pm$20.3 & 77.3$\pm$0.3 \\
        SGC       & 81.0$\pm$0.1 & 71.7$\pm$0.3 & 77.9$\pm$0.5 & OOM & OOM & 82.0$\pm$0.4 & 90.9$\pm$0.2 & 71.3$\pm$0.2 \\
        APPNP     & \textbf{\textcolor{red}{83.0$\pm$0.7}} & \textbf{\textcolor{red}{72.3$\pm$0.4}} & \textbf{\textcolor{red}{80.2$\pm$0.2}} & 92.5$\pm$0.2 & 93.3$\pm$0.2 & 83.2$\pm$0.7 & 91.7$\pm$0.7 & 77.6$\pm$0.3 \\
        CurvGN    & 82.3$\pm$0.5 & 71.9$\pm$0.6 & 78.9$\pm$0.4 & 92.5$\pm$0.3 & 93.4$\pm$0.2 & 83.5$\pm$0.5 & 91.3$\pm$0.5 & 75.4$\pm$0.3 \\
        \midrule
        CGNN\_Linear\_Sym      & 81.6$\pm$0.6 & 71.6$\pm$0.6 & 78.2$\pm$0.4 & 93.0$\pm$0.3 & 93.5$\pm$0.4 & 83.5$\pm$0.6 & 91.6$\pm$0.4 & 77.3$\pm$0.3 \\
        CGNN\_Linear\_$1^{st}$ & 81.6$\pm$0.6 & 71.5$\pm$0.5 & 78.0$\pm$0.3 & 92.5$\pm$0.3 & 93.4$\pm$0.3 & 83.5$\pm$0.6 & \textbf{91.8$\pm$0.5} & \textbf{\textcolor{red}{78.5$\pm$0.2}} \\
        CGNN\_Linear\_$2^{nd}$ & 81.5$\pm$0.5 & 71.8$\pm$0.6 & 78.1$\pm$0.4 & \textbf{\textcolor{red}{93.5$\pm$0.3}} & \textbf{\textcolor{red}{93.8$\pm$0.4}} & \textbf{\textcolor{red}{84.0$\pm$0.7}} & 91.5$\pm$0.7 & 76.3$\pm$0.4 \\
        CGNN\_Exp\_Sym         & 82.5$\pm$0.6 & 71.8$\pm$0.7 & \textbf{79.4$\pm$0.3} & 92.9$\pm$0.3 & 93.5$\pm$0.4 & 83.5$\pm$0.5 & 91.5$\pm$0.5 & 76.7$\pm$0.3 \\
        CGNN\_Exp\_$1^{st}$    & \textbf{82.8$\pm$0.7} & 71.4$\pm$1.0 & 78.4$\pm$0.4 & 92.1$\pm$0.2 & 93.5$\pm$0.4 & 83.5$\pm$0.6 & \textbf{\textcolor{red}{91.9$\pm$0.4}} & \textbf{78.3$\pm$0.2} \\
        CGNN\_Exp\_$2^{nd}$    & 82.5$\pm$0.6 & \textbf{72.1$\pm$0.7} & 78.9$\pm$0.5 & \textbf{93.2$\pm$0.3} & \textbf{93.7$\pm$0.3} & \textbf{83.8$\pm$0.7} & 91.4$\pm$0.6 & 75.7$\pm$0.4 \\
    \bottomrule
    \end{tabular}
    }
\end{table}

\subsection{Visual Analysis of Node Classification}

\begin{figure}
	\centering
		\includegraphics[width=0.95\textwidth]{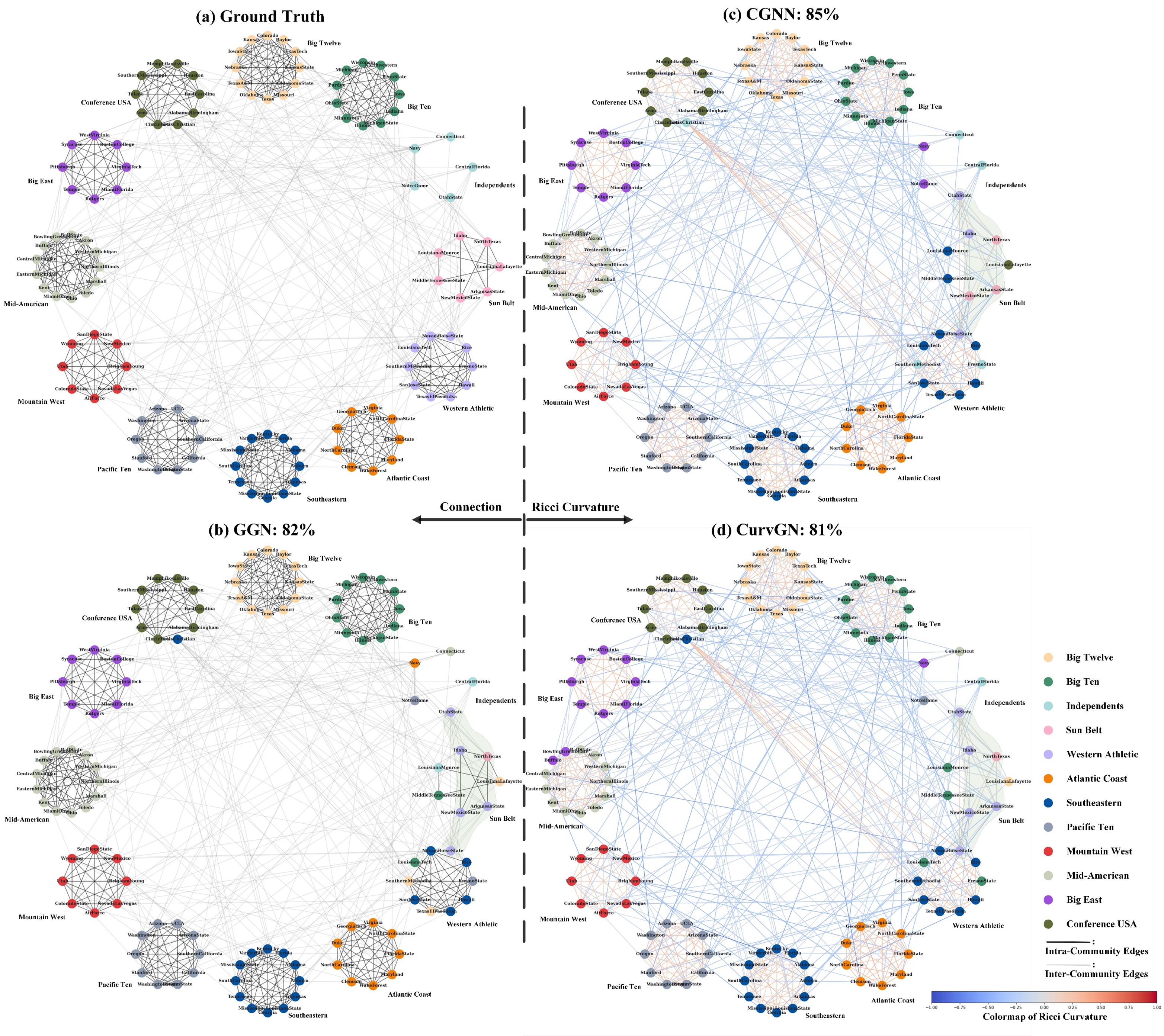}
	\caption{Visualization of the node classification prediction results of three typical models on the Football dataset. The colors of nodes indicate the predicted classes. The black edges of (a) and (c) represent the intra-community edges, while the grey edges represent the inter-community edges. The color of edges in (b) and (d) means the Ricci curvatures of the edges.}
	\label{fig: Football}
\end{figure}

In this section, we select a simple but community-structured dataset, the football dataset, to explore how the structural information affects the classification results of different GNNs in detail. The dataset is the schedule of Division I America college football games for the 2000 season: nodes indicate the teams of colleges and edges indicate regular-season games between two teams. The 12 college football conferences form the communities, and the nodes within the same community are considered to be of the same class. According to football college conference memberships, nodes are grouped together and color-coded, as shown in Figure \ref{fig: Football}(a). And the intra-community edges are relatively tight, while the inter-community edges are relatively sparse. Then, we compute the Ricci curvature of the edges and represent the Ricci curvature by color, as shown in Figure \ref{fig: Football}(c), (d). The curvature of the intra-community edges is almost always positive while that of the inter-community edges is mostly negative. We note that the positive curvature of inter-community edges is mainly located at the nodes TexasChristian and UtahState, due to the fact that most of the neighboring nodes of these two nodes belong to the communities Western Athletic and Sun Belt respectively.

We use the adjacency matrix of the dataset as node features, randomly select one node for each class as the label, and the rest of the nodes as the test set. The colors of the nodes in Figure \ref{fig: Football}(b), (c), and (d) represent the prediction result of GCN, CGNN, and CurvGN respectively. The classification accuracy of CGNN (85\%) outperforms GCN (82\%) and CGNN (81\%) which indicates that CGNN makes better use of the structural information. Further, we illustrate that the Ricci curvature efficiently improves the adaptive locality ability of GNNs by analyzing the nodes in the light green shaded area. Both GCN and CurvGN misclassified ArkansasState and NewMexicoState in Sun Belt as Western Athletic, while CGNN classified them correctly. The intersection of neighbor nodes of ArkansasState and NewMexicoState are: BoiseState, NorthTexas, Idaho, UtahState. Among them, BoiseState and NorthTexas are the labeled nodes of Western Athletic and Sun Belt. For the Idaho and UtahState nodes, all three models misclassify them into Western Athletic class.

\begin{figure}
	\centering
		\includegraphics[width=0.95\textwidth]{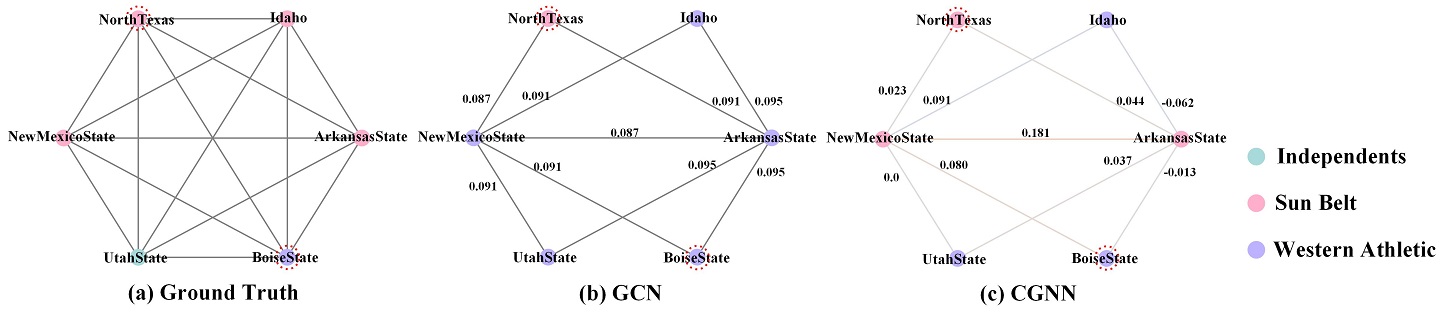}
	\caption{Visualization of the node classification prediction results of three typical models on the Football dataset. The colors of nodes indicate the predicted classes. The black edges of (a) and (c) represent the intra-community edges, while the grey edges represent the inter-community edges. The color of edges in (b) and (d) means the Ricci curvatures of the edges.}
	\label{fig: subgraph}
\end{figure}

The key to correctly predicting the ArkansasState and NewMexicoState classes is to weaken the impact of BoiseState, Idaho, and UtahState on these two nodes and strengthen that of NorthTexas. The six nodes coincidentally constitute a fully connected subgraph, as shown in Figure \ref{fig: subgraph}(a). We further simplify the subgraph to keep only the edges connecting ArkansasState and NewMexicoState. Since the node degrees of ArkansasState, NewMexicoState, NorthTexas, BoiseState, Idaho, and UtahState are 11, 12, 11, 10, 10, 10, GCN considers BoiseState, Idaho, and UtahState play a more important role for ArkansasState and NewMexicoState than NorthTexas as shown in Figure \ref{fig: subgraph}(b). However, the curvature of NorthTexas to ArkansasState and NewMexicoState is significantly greater than that of BoiseState, Idaho, and UtahState. And CGNN properly utilizes this property to weaken the effect of BoiseState, Idaho, and UtahState and to strengthen that of NorthTexas, as shown in Figure \ref{fig: subgraph}(c). CurvGN does not utilize the Ricci curvature, and also misclassifies BowlingGreenState and Buffalo in Mid-American as Big East classes. The result suggests that CGNN effectively exploits the property of Ricci curvature to enhance the ability to evaluate the strength of connections between nodes on local structures.

\subsection{Ablation Experiment}

\begin{figure}
	\centering
		\includegraphics[width=0.55\textwidth]{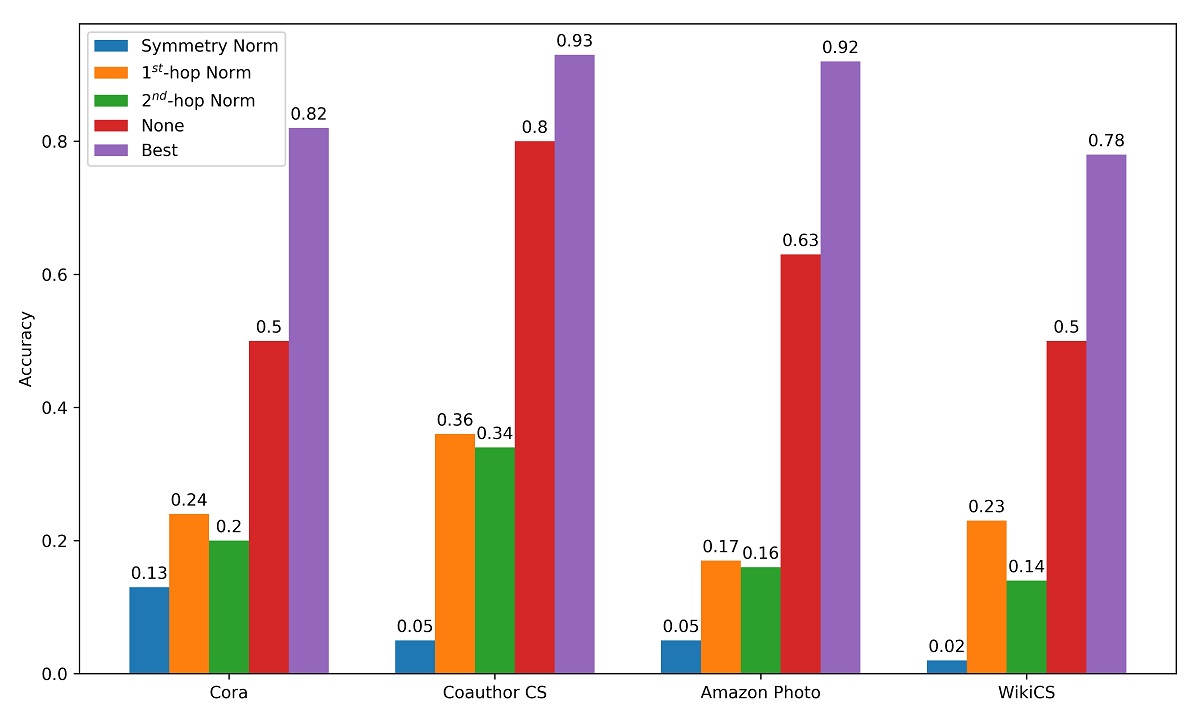}
	\caption{Comparison of predicted accuracies under difference normalization ways with/without NCTM. Best represents the best accuracies of CGNN. None indicates CGNN without NCTM nor CNM. The rest indicates CGNN without NCTM.}
	\label{fig: ablation_NCTM}
\end{figure}

We design ablation experiments to illustrate the necessity of NCTM. Firstly, we compare the accuracy of CGNN without NCTM to the best accuracy achievable by CGNN on four datasets, and the results are shown in Figure \ref{fig: ablation_NCTM}. We find that once NCTM is removed, the performance of CGNN inevitably degrades significantly. In this condition, CGNN with CNM performs even worse than the one without CNM. Due to the negative curvature, the symmetry normalization may generate imaginary weights of node features, resulting in worse accuracy than that of random ones. For $1^{st}$-hop normalization or $2^{nd}$-hop normalization, the performance of the model is also only slightly better than that of the random one. This experiment shows that NCTM has a crucial impact on the convergence of CGNN.

\begin{figure}
	\centering
		\includegraphics[width=0.55\textwidth]{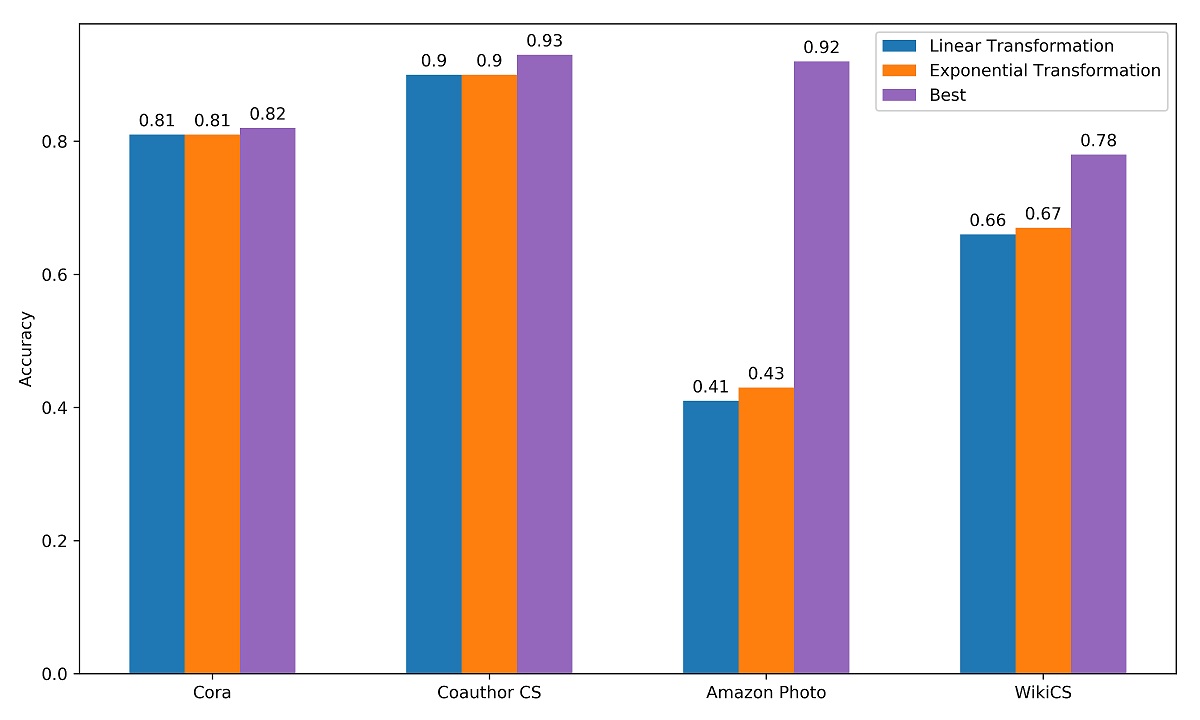}
	\caption{Comparison of predicted accuracies under difference negative curvature transformation ways with/without CNM. Best represents the best accuracies of CGNN, and the rest indicates CGNN without CNM.}
	\label{fig: ablation_CNM}
\end{figure}

Further, we explored the effect of CNM on the adaptation of CGNN to various datasets. We compare the accuracy of CGNN without CNM to its best performance, and the results are shown in Figure \ref{fig: ablation_CNM}. The accuracy of CGNN without CNM only decreases by 1 to 3 percentage points on Cora and Coauthor CS, but it decreases by more than 10 percentage points on the other two datasets. In particular, on Amazon Photo, the accuracy drops by 50 percentage points. It is worth noting that Cora and Coauthor CS are both citation-based datasets, while Amazon Photo is an e-commerce dataset and WikiCS is an Internet dataset. These results show that the normalization module can effectively extend the range of datasets to which CGNN is adapted.

\section{Conclusion}
In this paper, we propose a novel graph neural network model called CGNN, improving the discriminative power on the structural importance of neighboring nodes by exploiting Ollivier's Ricci curvature. CGNN utilizes the structural signatures of the Ricci curvature which measure the interaction or overlap between the neighborhoods of pairwise nodes, to selectively aggregate the features of neighboring nodes. For stabilizing the training and improving the performance of CGNN, we transform the negative curvature into positive by the NCTM and normalize the Ricci curvature by the CNM according to the structural information of different order hops. Experimental results on synthetic datasets show that CGNN significantly outperforms baselines when community structures exist in datasets. For the real-world datasets with heterogeneous topology, CGNN achieves comparable or even better results than baselines. Further, we illustrate that CGNN can enhance the connections between intra-community nodes and weaken the connections between inter-community nodes by analyzing the classification result of CGNN, GCN and CurvGN in detail on the football dataset. The ablation experiments show that the NCTM and the CNM are crucial for the superior performance of GCNN.

In the future, we will explore how the Ricci curvature relieves the oversmoothing of GNNs and improves the generalization ability of GNNs. Randomly dropping fractional edges during the training of GNNs has been demonstrated to be an effective way to alleviate the oversmoothing triggered by stacking lots of layers \cite{rong2019dropedge}. For the node classification task, only oversmoothing node representations of different classes is harmful, while smoothing node representations of the same class is beneficial. Inspired by Ricci curvature used for community detection, we can drop edges based on Ricci curvature, e.g., edges with negative curvature have a higher probability of being deleted than those with positive curvature. We also try to extend the Ricci curvature to graph classification tasks, because the Ricci curvature represents the connectivity of the local structure rather than a specific graph. We consider that the local structures are similar for the same class of graphs, and the Ricci curvature distributions are also similar. The Ricci curvature combined with the learning ability of GNNs may empower models to generate more representative graph representations, and improve the generalization of models to unknown datasets.

\bibliographystyle{unsrt}
\bibliography{CGNN-refs}

\end{document}